\documentclass[10pt,twocolumn,letterpaper]{article}
\usepackage[pagenumbers]{cvpr} 
\usepackage{graphicx}
\usepackage{amsmath}
\usepackage{amssymb}
\usepackage{booktabs}

\usepackage[pagebackref,breaklinks,colorlinks]{hyperref}
\usepackage[capitalize]{cleveref}
\crefname{section}{Sec.}{Secs.}
\Crefname{section}{Section}{Sections}
\Crefname{table}{Table}{Tables}
\crefname{table}{Tab.}{Tabs.}
\def\cvprPaperID{5645} 
\def\confName{ICCV}
\def\confYear{2023}
\begin{document}
\title{Internal-External Boundary Attention Fusion for Glass Surface Segmentation}
\author{
Dongshen Han$^{1}$ \qquad 
Heechan Yoon$^{1}$ \qquad 
Hyukmin Kwon$^{2}$ \qquad 
Hyun-Cheol Kim$^{2}$ \qquad \\
Hyon-Gon Choo$^{2}$ \qquad 
Seungkyu Lee$^{1}$  \qquad 
Chaoning Zhang$^{1}$ \qquad 
\\
$^1$Kyunghee University \\
$^2$Electronics and Telecommunications Research Institute
\\
\tt\small \ seungkyu@khu.ac.kr  chaoningzhang1990@gmail.com}


\maketitle

\begin{abstract}
Glass surfaces of transparent objects and mirrors are not able to be uniquely and explicitly characterized by their visual appearances because they contain the visual appearance of other reflected or transmitted surfaces as well. Detecting glass regions from a single-color image is a challenging task. Recent deep-learning approaches have paid attention to the description of glass surface boundary where the transition of visual appearances between glass and non-glass surfaces are observed. In this work, we analytically investigate how glass surface boundary helps to characterize glass objects. Inspired by prior semantic segmentation approaches with challenging image types such as X-ray or CT scans, we propose separated internal-external boundary attention modules that individually learn and selectively integrate visual characteristics of the inside and outside region of glass surface from a single color image. Our proposed method is evaluated on six public benchmarks comparing with state-of-the-art methods showing promising results. 
\end{abstract}
\vspace*{-4mm}
\section{Introduction}
\label{sec:intro}
Transparent objects and mirror surfaces are everywhere around such as windows, bottles, eye glasses and omnipresent mirrors. They are expected to be detected, localized and reconstructed from color images in computer vision tasks as other opaque objects are done. However, glass surface is not able to be uniquely characterized by their visual appearances. Most transparent surface region show visual appearance of both transmitted background scene and reflected objects. Visual appearance observed from mirror surface is entirely from reflected objects. Therefore, visual characteristics of glass object are not able to be uniquely defined and observed. This makes performance limitations of computer vision methods in practical many applications. For example, autonomous mobile robot may collide with transparent front door or mirror wall. Robot arm struggles to grip a transparent bottle. 
\begin{figure}
  \centering
\includegraphics[scale=0.3]{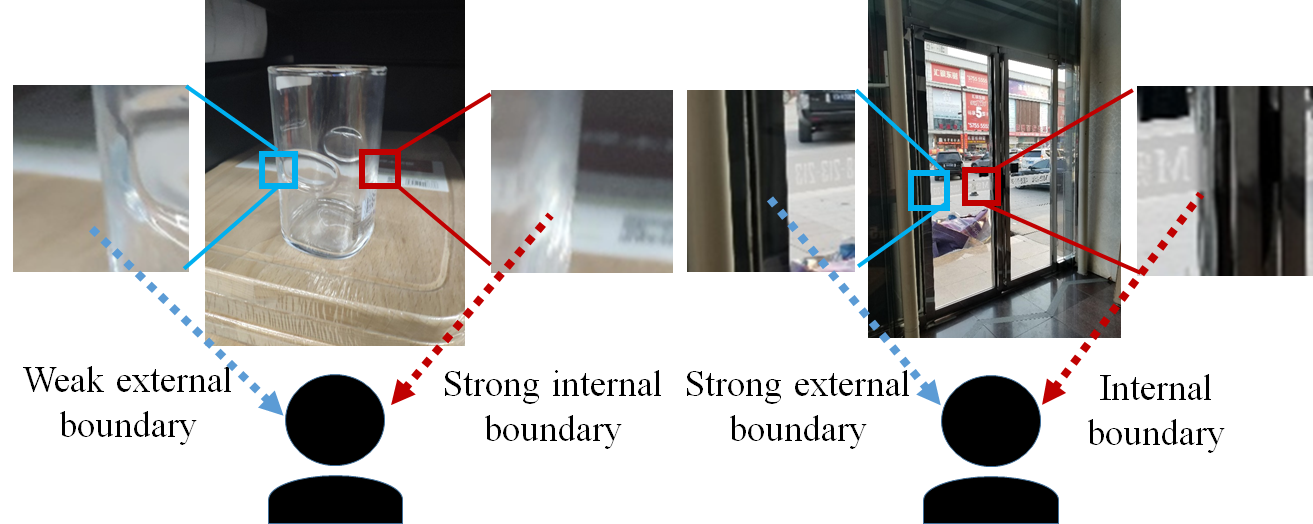}
\caption{Glass objects: Glass cup with strong internal but weak external boundaries and Window with strong external boundary}
\vspace*{-4mm}
  \label{firstpic}
\end{figure}
Semantic segmentation methods using convolutional neural networks \cite{zhang2023understanding} have tried to identify visual texture and corresponding surface material type of objects \cite{long2015fully} \cite{ronneberger2015u} \cite{badrinarayanan2017segnet} \cite{zhang2023complete} \cite{zhang2023one}. Long \etal~\cite{long2015fully} transpose feature extraction layers to a segmentation task and update parameters by fine tuning. At the same time, it designs a novel structure that combines low-level and high-level semantics, which is extensively utilized by following semantic segmentation networks such as \cite{zhao2017pyramid} .
Lin \etal~\cite{lin2019refinenet} integrate the features of encoder and decoder phases and utilizes chained residual pooling in the module to capture contextual information over a wide range of contexts. Non-local operators \cite{wang2018non} based on a self-attention mechanism \cite{chorowski2015attention} are applied in semantic segmentation \cite{fu2019dual}\cite{babiloni2020tesa}\cite{zhang2023faster}. Song \etal~\cite{song2022fully} propose fully attentional blocks to encode spatial and channel attentions in similarity graphs with high computational efficiency. Most of these methods, however, fail to characterize common visual appearance of glass surface within the boundary of transparent or mirror object region.

Recently, glass surface object segmentation methods are proposed including transparent objects \cite{xie2020segmenting} \cite{zheng2022glassnet} \cite{he2021enhanced} \cite{he2021enhanced} and strongly reflective objects such as mirror \cite{guan2022learning} \cite{mei2021depth} \cite{yang2019my} \cite{han2023segment}. Mei \etal~\cite{mei2020don} propose to use different dilation convolutions and thus use different sensory fields to capture contrast features and discontinuities in semantic information. However, overlapped non-glass objects lead to semantic discontinuities.
Xie \etal~\cite{xie2020segmenting} propose a new benchmark for glass object segmentation and considers the characteristics of transparent object boundary features. EBLNet \cite{he2021enhanced} extracts boundary features employing a non-local approach. It enhances whole boundary features of glass surface and consider the relationship between the inside and the boundary of glass object region. By assigning additional attention on the boundary of glass object, they alleviate the problem of the deficiency of common visual appearance inside the boundary of transparent or mirror objects. Guan \etal~\cite{guan2022learning} propose a model to exploit the semantic associations between the mirror and its surrounding objects for a reliable mirror localization. However, the method tends to be biased to the surrounding objects appeared in the training dataset that are not core features of glass objects. Xie \etal~\cite{xie2021segmenting} employ attentional mechanisms to explore relationships among glass objects of different categories. Reflections and ghosts are employed as clues for the identification of glass surface. RCAM\cite{lin2021rich} guides the network to identify reflections based on the feature information obtained from the reflection removal network\cite{zhang2018single}.


On the other hand, external boundaries have played a crucial role in salient object segmentation\cite{wei2020f3net}\cite{zhu2023sharp}\cite{qiao2023robustness}, especially in medical imaging, such as CT image segmentation\cite{ghadimi2015skull} \cite{shojaii2005automatic} \cite{bai2016automatic}. In CT scan or X-ray images, visual textures of human organ are not clearly separable to each other. Instead, thick skeleton or outer boundary features have been utilized for the segmentation of organ regions in CT images\cite{shojaii2005automatic}. 
In CT scans, boundary of organ region is ambiguous containing lots of noises. In X-ray images, bones and organs are overlapped showing confused boundary connections. In such cases, organ region segmentation focuses on the boundary and texture features of external boundary, which is the order of human observation\cite{ghadimi2015skull}.
Our careful observations on the prior methods and diverse transparent and mirror objects bring the conclusion that neither reflections or ghosts inside glass region nor surroundings of glass object can reliably describe glass surface. As the clues (either at inside or outside of glass region) are located farther from the boundary of a glass object, observed features are less relevant to the characteristics of glass surface. Instead, we have to focus more on the inner and outer vicinity of the boundary. Outer vicinity of the boundary contains coherent appearance that decide the physical frame of a glass object. Inner vicinity strongly reveals the characteristics of glass surface based on the light refraction or distortion, while is less affected by the appearance of transmitted background objects than the middle area of glass surface. These two aspects compositely describe the existence of glass surface. Actually, this scheme is how human perceive glass surface under challenging environment.

According to the insight, we define external boundary (exboundary) and internal boundary (inboundary) as respective region of outer vicinity and inner vicinity of the boundary of glass object. Our aim is to address glass object surface segmentation problem. We design deep neural networks based on the following four observations: (1) Human recognize an object from salient features first. This is also obvious with glass objects. Human tend to identify the potential area of a glass surface object from its easily distinguishable internal and external boundary regions, rather than entire glass surface region. (2) Glass surface objects show diverse internal and external boundary appearances and their relative importance. Window frame is strong clue observed from external boundary region of transparent surface. A glass bottle does not have external boundary but have strong internal boundary characteristics from light refraction. (3) Internal and external boundaries have different role is the recognition of glass surface region. Window frame observed at the external boundary of window decide potential glass surface region. Further investigation of internal boundary searching for reflections or ghosts determines the presence of window glass. Ignoring such semantic differences between internal and external boundary regions may result in incorrect segmentation of window frames without window glass or opened doors. (4) Closer region to the boundary shows stronger characteristics of glass surface. Figure \ref{firstpic} shows two challenging glass objects. Glass cup shows strong internal boundary characteristics but weak external boundary. On the other hand window has strong external boundary clue (window frame).

Based on our study and observations we claim that glass object segmentation networks have to collect external and internal boundary features individually. And then they have to be selectively integrated for potential glass surface region segmentation and presence determination of glass surface, respectively. To this end, we construct Internal-External Boundary Attention Module (IEBAM) with step-by-step supervision improving the feature fusion of residual networks\cite{he2016deep}.
IEBAM utilizes an attention mechanism to implement the human scheme of glass surface detection. It extracts salient external and internal boundary features of glass surface object such as glass frame and light distortion respectively. 
After that it obtains more detailed internal boundary features for glass detection. In order to handle glass surface objects with weak external boundary such as glass cups where external boundary fails to provide significant features, we propose Fused Boundary Attention Module (FBAM). FBAM combines internal and external boundaries and weighs more on internal boundary features by learning the semantic clue of transparent objects. Glass surface region is determined by finding the relationship between the reinforced boundary and glass features such as reflections occurring inside the transparent object.

\begin{figure*}[t!]
\centering
\begin{subfigure}{\linewidth}
\centering
\includegraphics[scale=0.26]{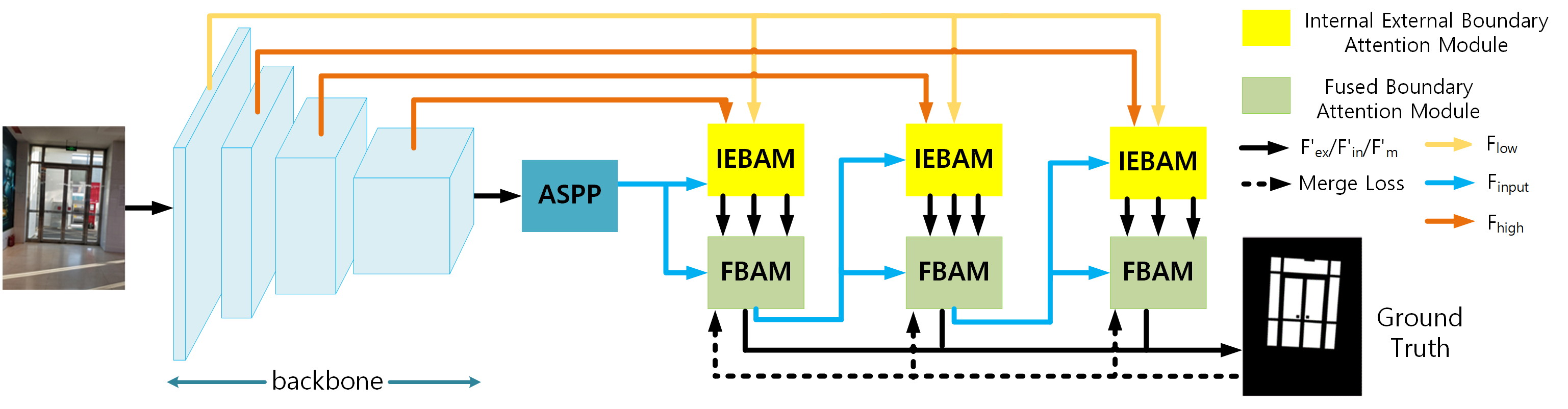}
\end{subfigure}
\caption{The pipeline of our proposed network. It can be observed that with the ASPP structure, our baseline is DeeplabV3+, and we connect our IEBAM and FBAM in a cascaded order.}
\label{fig:2}
\end{figure*}
\section{Methods}
\label{sec:formatting}
Our glass surface segmentation network (Figure \ref{fig:2}) is based on two newly proposed deep neural networks: Internal External Boundary Attention Module (IEBAM) and Fused Boundary Attention Module (FBAM). IEBAM separately learns internal and external boundary features, and exploits the internal boundary features to obtain the potential locations of glass surface object regions. FBAM learns glass surface object features from the previous network and fuses internal and external boundary features in proportion obtaining reinforced boundary features. By learning the relationship between features such as reinforced boundaries and reflections from glass regions, the confidence of glass regions are improved and non-glass regions are suppressed. Finally, inspired by contour loss \cite{chen2020contour}, we incorporate internal and external losses of spatial weight map assignment decomposition in cross-entropy loss.

Our framework is inspired by EBLNet\cite{he2021enhanced} as shown in Figure \ref{fig:2}. Backbone is DeeplabV3+\cite{chen2018reverse} and outputs multi-level features (layer 1, layer 2, layer 3, layer 4, and ASPP\cite{chen2018reverse}). Multi-level features are fed to Internal External Boundary Attention Module (IEBAM). IEBAM outputs internal, external boundaries and body features of glass surface. Fused Boundary Attention Module (FBAM) utilizes the output of IEBAM to refine the final transparent and glass region features. Entire network calculates losses and performs training in five different aspects: boundary-wise, external boundary-wise, internal boundary-wise, body-wise (non-boundary region), and segmentation-wise. Note that our proposed method shows significant advances over EBLNet in the following points. First, EBLNet collects features from entire boundary region all together. In our method, diverse inside and outside characteristics of transparent surface are separately identified and optimally utilized resulting in cleaner segmentation results. In PGM of EBLNet, using whole boundary areas containing non-glass region misleads glass detection. 

\subsection{Internal External Boundary Attention Module}
\begin{figure}[!h]
\centering
\begin{subfigure}[t]{1\linewidth}\includegraphics[width=1\linewidth]{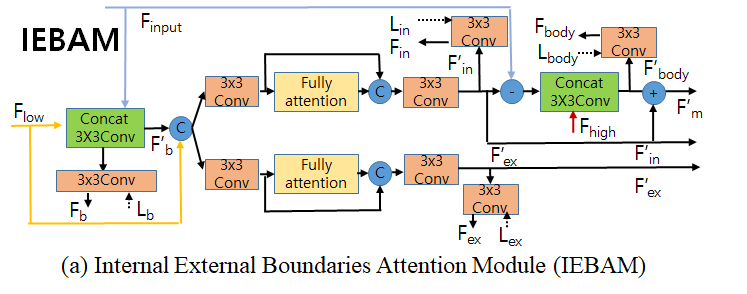}\end{subfigure}
\begin{subfigure}[t]{1\linewidth}\includegraphics[width=1\linewidth]{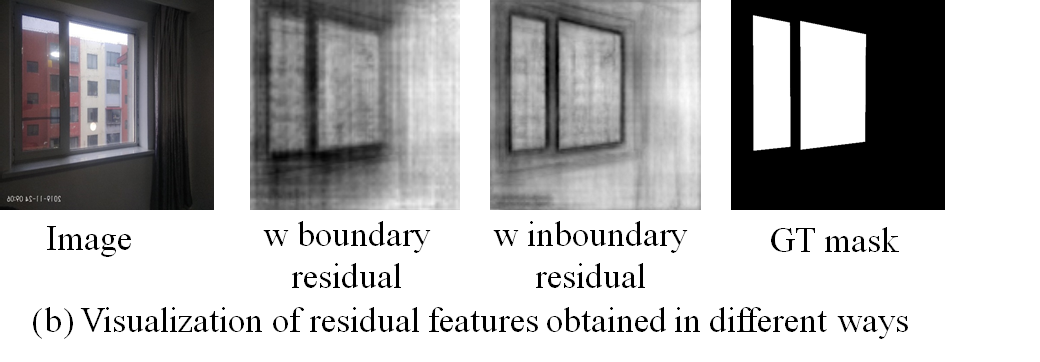}\end{subfigure}\caption{(a) IEBAM: ${\textit{L}}_{ex}$, ${\textit{L}}_{in}$, ${\textit{L}}_{b}$, ${\textit{L}}_{body}$ are loss functions (b) Vislualized Residual Features: Second and third images are PCA visualizations of residual features with entire boundary and internal boundary features, respectively. Using only internal boundary reduces noise compared to entire boundary.}
\label{IEBAM}
\vspace*{-3mm}
\end{figure}
IEBAM structure is shown in Figure \ref{IEBAM}(a). Input of an IEBAM module is the outputs ${\textit{F}}_{input}$, ${\textit{F}}_{high}$ and ${\textit{F}}_{low}$ generated by backbone network of each layer, thus obtaining feature maps of different receptive field of glass surface: (1) boundary feature maps, ${\textit{F}{'}}_{b}$, (2) internal boundary feature maps, ${\textit{F}{'}}_{in}$, (3) external boundary prediction, ${\textit{F}{'}}_{ex}$, (4) ‘body’ feature maps, ${\textit{F}{'}}_{body}$ (segmentation features except the boundary region), (5) ‘complete’segmentation feature maps, ${\textit{F}{'}}_{m}$ (that characterizes both body region and internal boundary region). Among them, ${\textit{F}{'}}_{ex}$, ${\textit{F}{'}}_{in}$ and ${\textit{F}{'}}_{m}$ are fed to the subsequent FBAM module. ${\textit{F}{'}}_{ex}$, ${\textit{F}{'}}_{in}$, ${\textit{F}{'}}_{b}$ and  ${\textit{F}{'}}_{m}$ employ convolution operation to obtain ${\textit{F}}_{ex}$, ${\textit{F}}_{in}$, ${\textit{F}}_{b}$ and ${\textit{F}}_{m}$, and perform loss calculations.

IEBAM is designed according to the observation scheme of human with glass surface object. Most easily identifiable internal and external boundary features are extracted first, and these are used as the basis for obtaining  internal and final glass surface features. Because lower layers of backbone networks contain cleaner boundary information, we concatenate ${\textit{F}}_{input}$ and the refine features ${\textit{F}}_{low}$ and obtain rough boundary features ${\textit{F}}_{b}$ by convolution operation. In order to obtain the internal boundary features ${\textit{F}}{'}_{in}$ accurately, we concatenate whole boundary region features and ${\textit{F}}{'}_{low}$. Fully attention network focuses on the correlation between internal boundary and ${\textit{F}}_{low}$ to obtain internal boundary attention. Finally, internal boundary features are refined based on the attention with convolution operations, which is same for external boundary defined as follows.
\begin{equation}
\label{eqn:01}
{\textit{F}}{'}_{in}={\textit{g}}_{3\times{3}}({\textit{f}}({\textit{g}}_{3\times{3}}([{\textit{F}}{'}_{b};{\textit{F}}_{low})]);{\textit{g}}_{3\times{3}}([{\textit{F}}{'}_{b};{\textit{F}}_{low}])),
\end{equation}
where [; ] and ${\textit{g}}_{3\times{3}}$ denote concatenation and convolution. ${\textit{f}}$ is fully attention network. ${\textit{F}}_{input}$ contains glass region features computed by the previous layer of the network. Different from EBLNet, where the whole boundary of the glass surface object features are subtracted from the ${\textit{F}}_{input}$, we subtract ${\textit{F}}{'}_{in}$ from the ${\textit{F}}_{input}$ to reduce the disturbance from the external boundary region. 
${\textit{F}}{'}_{high}$ contains a large amount of high receptive field information, so the glass body feature is concated with it and the final glass surface object body feature ${\textit{F}}{'}_{body}$ is obtained by convolution operation. This procedure is formulated as follows.
\begin{equation}
\label{eqn:02}
{\textit{F}}{'}_{body}={\textit{g}}_{3\times{3}}([{\textit{F}}{'}_{high};{(\textit{F}}_{input}-{\textit{F}}{'}_{in})]),
\end{equation}

Different from EBLNet which directly fuses the whole boundary features, we perform ${\textit{F}}{'}_{m}={\textit{F}}{'}_{body}+{\textit{F}}{'}_{in}$ thus avoiding external boundary region features that do not belong to the glass region and excluded from the final merged feature ${\textit{F}}{'}_{m}$ calculation. Finally, ${\textit{F}}{'}_{in}$, ${\textit{F}}{'}_{ex}$, ${\textit{F}}{'}_{body}$, ${\textit{F}}{'}_{m}$ are passed through the convolution operation and loss calculation separately.
We compare generated feature map ${\textit{F}}{'}_{body}$ in the process of IEBAM with EBLNet, which are formulated as follows. 
\begin{equation}
\label{eqn:02}
{\textit{F}}{'}_{resiudal}={\textit{F}}_{input}-{\textit{F}}{'}_{in},
\end{equation}
\begin{equation}
\label{eqn:02}
{\textit{F}}{'}_{resiudal}={\textit{F}}_{input}-{\textit{F}}{'}_{b},
\end{equation}
where ${\textit{F}}{'}_{resiudal}$ is the residual feature map generated in the process of obtaining ${\textit{F}}{'}_{body}$, and ${\textit{F}}{'}_{residual}$ is visualized by dimensionality reduction using principal component analysis (PCA) in Figure \ref{IEBAM}(b). Employing ${\textit{F}}{'}_{in}$ reduces noise compared to employing ${\textit{F}}{'}_{b}$.

\subsection{Fused Boundary Attention Module}
\begin{figure}[!b]
\vspace*{-3mm}
\centering
\begin{minipage}[t]{1\linewidth}\includegraphics[width=1\linewidth]{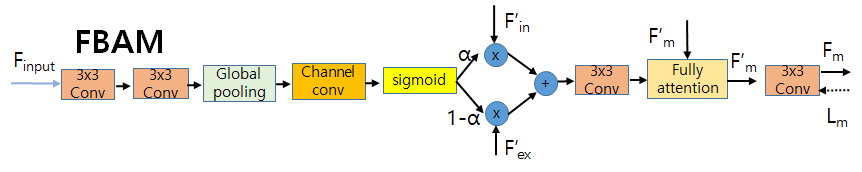}\end{minipage}%
\caption{The structure of our proposed FBAM. ${\textit{L}}_{m}$ is the loss of entire glass prediction. }
\label{fig:4}
\end{figure}
FBAM structure is shown in Figure \ref{fig:4}. Although internal and external boundaries of glass surface containing different features are obtained, boundary features vary over different glass surface types. Windows and mirrors usually contain strong external boundaries, while glass cups and bottles have very weak external boundaries. Therefore, we fuse internal and external boundary features based on the semantic information of glass surface objects extracted from the previous module. We are inspired in the fusion of depth image features and RGB image features \cite{mei2021depth}.
We use sigmoid to assign weights to internal and external boundaries. Ground truth of our internal and external boundary regions share one pixel thickness of real boundary as shown in Figure \ref{gth detail}(b). So we discard Relu and other nonlinear functions with outputs equal to zero. In particular, to fuse internal and external boundary features proportionally, we perform global pooling and convolution operations on the glass segmentation features obtained from previous network, and obtain boundary attention after employing sigmoid function. Internal and external boundary features are fused proportionally by the boundary attention values, and the boundary-enhancing feature map ${\textit{F}}{'}_{en}$ are obtained by convolution operation, which can be formulated as follows.
\begin{equation}
\label{eqn:06}
\mu(\textit{F})={\textit{g}}_{1\times{1}}(\textit{R}({\textit{N}}({\textit{g}}_{1\times{1}}({\textit{G}_{3\times{3}}}(\textit{F}_{input}))))),
\end{equation}
\begin{equation}
\label{eqn:06}
{\textit{G}_{3\times{3}}}(\textit{F}_{input})=G({\textit{g}}_{3\times{3}}({\textit{g}}_{3\times{3}}(\textit{F}_{input}))),
\end{equation}
\begin{equation}
\label{eqn:07}
\alpha=\frac{1}{1+\exp(-\mu({\textit{F}}{'}_{in}))  },
\end{equation}
\begin{equation}
\label{eqn:07}
\beta=1-\alpha,
\end{equation}
\begin{equation}
\label{eqn:07}
{\textit{F}}_{en}={\textit{g}}_{3\times{3}}(\alpha\cdot{\textit{F}}{'}_{in}+\beta\cdot{\textit{F}}{'}_{ex}),
\end{equation}
where ${\textit{G}}$ is global average pooling operation. ${\textit{R}}$, ${\textit{N}}$ are ReLU and Batch Normalization (BN) activation function. 
Both enhanced boundary features ${\textit{F}}{'}_{en}$ and unique glass surface object features (ghosts, reflections) ${\textit{F}}{'}_{m}$ belong to the same semantic of glass object revealing different characteristics. We use two convolution operations to transfer the enhanced boundary features to the feature space of glass surface object body. Then, we use ${\textit{F}}{'}_{en}$ as query and utilize the fully-attention method to improve the score of glass surface object merged regions.
Specially, We utilize ${\textit{F}}{'}_{en}$ as a fully-attention query feature, thus the relationship between features such as reflections and ghosts specific to the glsss region ${\textit{F}}{'}_{m}$ and ${\textit{F}}{'}_{en}$ are detected so that ${\textit{F}}{'}_{m}$ can be refined as formulated below.
\begin{equation}
\label{eqn:03}
{\textit{F}}{'}_{m}=\gamma\sum_{i=1}^{c}\frac{\exp( {\textit{Q}}_{en(i)} \cdot{\textit{K}}_{m(j)}) }{\sum_{i=1}^{c}{\exp( {\textit{Q}}_{en(i)} \cdot{\textit{K}}_{m(j)}) }} \cdot{\textit{V}}_{m(j)} \\\
+{\textit{F}}{'}_{m},
\end{equation}

\subsection{Decomposed Contour Loss}
\begin{figure}[!b]
\centering
   \begin{subfigure}[t]{0.3\linewidth}\includegraphics[width=0.9\linewidth]{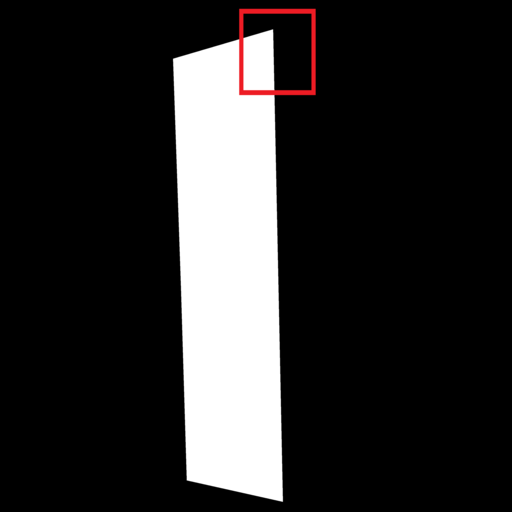}\caption{Ground Truth}\end{subfigure}
    \begin{subfigure}[t]{0.67\linewidth}\includegraphics[width=0.9\linewidth]{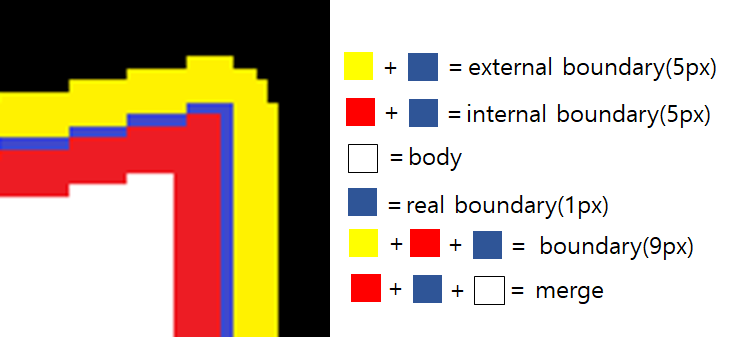}\caption{Ground Truth detail}\end{subfigure}
    \caption{(a) Ground truth of a glass surface object (b) Internal boundary, external boundary, real boundary, body and merged regions of glass object with ground truth distribution}
    \label{gth detail}
\end{figure}

The location of internal and external boundaries are ambiguous and the thickness of useful boundary region such as glass frame is not constant.
Inspired by contour loss\cite{chen2020contour}, we apply spatial weight maps in cross entropy loss, which assigns relatively high value to emphasize pixels near the external and internal glass boundaries. External boundary spatial weight map ${\textit{M}}^{C}_{ex}$ is formulated as follows.
\begin{equation}
\label{eqn:11}
{\textit{M}}^{C}_{ex}={\textit{g}}^{th}_{ex}\cdot{Guasss}({\textit{g}}^{th}_{b})+1,
\end{equation}
${\textit{g}}^{th}_{b}$, ${\textit{g}}^{th}_{ex}$ are the ground truth of boundary and external boundary region. In our experiments, thickness of ${\textit{g}}^{th}_{b}$ is set to 9 and ${\textit{g}}^{th}_{in}$, ${\textit{g}}^{th}_{ex}$ are 5 respectively. Internal and external boundary include real boundary of thickness 1, as shown in Figure \ref{gth detail}(b). And then, Gaussian smoothing filter is applied to obtain spatial weight map of the whole boundary.
Contour loss ${\textit{L}}_{in}$ is implemented as following formula.
\begin{multline}
\label{eqn:05}
{\textit{L}}_{in}=-\sum_{x,y}{\textit{M}}^{C}_{in(x,y)}\cdot({\textit{Y}}_{in(x,y)}\cdot{\log{\textit{g}}^{th}_{in(x,y)}}\\
+(1-{\textit{Y}}_{in(x,y)})\cdot\log{(1-{\textit{g}}^{th}_{in(x,y)})}),
\end{multline}

${\textit{Y}}_{in(x,y)}$ indicates predicted saliency map and the loss of external boundary is calculated in the same way. 
\subsection{Network Architecture and Loss Function}
Proposed modules are trained end-to-end way for glass object bodies and glass surface boundaries respectively. IEBAM outputs ${\textit{F}}_{in}$, ${\textit{F}}_{ex}$, ${\textit{F}}_{b}$, ${\textit{F}}_{body}$ which are the predictions of the internal boundary, external boundary, boundary and glass object body respectively. FBAM outputs ${\textit{F}}_{m}$ which is the prediction of final integrated glass surface object region. Joint loss function is as follows.
\begin{multline}
\label{eqn:05}
{\textit{L}}_{joint}={\textit{L}}_{b}({\textit{F}}_{b},{\textit{G}}_{b})+{\textit{L}}_{in}({\textit{F}}_{in},{\textit{G}}_{in})+{\textit{L}}_{ex}({\textit{F}}_{ex},{\textit{G}}_{ex})\\ +{\textit{L}}_{body}({\textit{F}}_{body},{\textit{G}}_{body})+{\textit{L}}_{m}({\textit{F}}_{m},{\textit{G}}_{m}),
\end{multline}
${\textit{G}}_{m}$, ${\textit{G}}_{b}$, ${\textit{G}}_{in}$, ${\textit{G}}_{ex}$ and ${\textit{G}}_{body}$ indicate original boundary, internal boundary, external boundary, and ground truth of body as shown in Figure \ref{gth detail}(b). ${\textit{L}}_{b}$ is a normal Dice Loss\cite{milletari2016v}, while ${\textit{L}}_{in}$, ${\textit{L}}_{ex}$ adopt decomposed contour loss. ${\textit{L}}_{m}$ and ${\textit{L}}_{body}$ are standard Cross-Entropy Loss.

\section{Experimental Evaluations}
\textbf{Datasets:}
We conduct experiments on four glass segmentation and two mirror segmentation datasets (Trans10k \cite{xie2020segmenting}, GDD\cite{mei2020don}, GSD\cite{lin2021rich}, GSD-S\cite{linexploiting} and MSD\cite{yang2019my}, PMD\cite{lin2020progressive}). GDD contains 2980 training and 936 test images. Trans10K is the largest glass object segmentation dataset with 5000 training images, 1000 validation images and 4428 test images. It has two categories: stuff and things. GSD and GSD-S are the latest transparent object datasets collected and labeled through networking and photography. GSD consists of 4102 annotated glass images, and contains close up, medium and long shots from diverse scenes. The GSD-S dataset includes semantic information about other objects in the images, however, we do not utilize it. MSD is a large mirror segmentation dataset with 4018 images (3063 training, 955 test) and PMD is the latest mirror segmentation dataset collected in outdoor and indoor. 

\textbf{Implementation Details:}
Our implementation is based on PyTorch, and our model is trained and tested on a PC with an 16-core i7-9700K 3.6 GHZ CPU, and 8 NVIDIA GeForce RTX 3090 GPU cards, including Trans10k, GDD, and MSD datasets. We are not using conditional random fields (CRF) \cite{fields2001probabilistic} as a post-processing.

\textbf{Evaluation Metrics:}
We use five evaluation metrics commonly used in sementic and glass surface segmentation tasks: intersection over union (IoU), pixel accuracy (ACC), weighted F-measure (\&${F}_{\beta}$) \cite{margolin2014evaluate}, mean absolute error (MAE), and balance error rate (BER). Since Trans10k has two categories: stuff and things, we adopt mean intersection over union (mIou), mean balance error rate (mBer) as the evaluation metrics. 
${F}_{\beta}$ is a harmonic mean of average precision and average recall defined as follows.
\begin{equation}
\label{eqn:03}
{\textit{F}_{\beta}}=\frac{(1+{\beta^{2}})(Precision\times{Recall})}{{\beta^{2}}Precision+Recall},
\end{equation}
where $\beta^{2}$ is set to 0.3 as suggested in \cite{achanta2009frequency}.
Mean absolute error (MAE) is widely used in foreground-background segmentation tasks where average pixel-wise error between predicted mask P and ground truth mask G are calculated.
\begin{equation}
\label{eqn:03}
{\textit{MAE}}=\frac{1}{H\times{W}}\sum_{i=1}^{H}\sum_{j=1}^{W}\mid P(i,j)-G(i,j)\mid,
\end{equation}
where P(i, j) indicates predicted probability at location (i, j).
We employ balance error rate (BER/mBER), which takes into account the imbalance region in transparent (mirror) and non-transparent (non-mirror) regions quantitatively evaluating the performance of glass surface segmentation.
\begin{equation}
\label{eqn:03}
{\textit{BER}}=(1-\frac{1}{2}(\frac{TP}{\textit{N}_{p}}+\frac{TN}{\textit{N}_{n}}))\times{100},
\end{equation}
where TP, TN, Np, and Nn represent the numbers of true positives, true negatives, glass pixels, and non-glass pixels.
\subsection{Experiment on Trans10k}
\begin{figure*}
\centering
\includegraphics[scale=0.8]{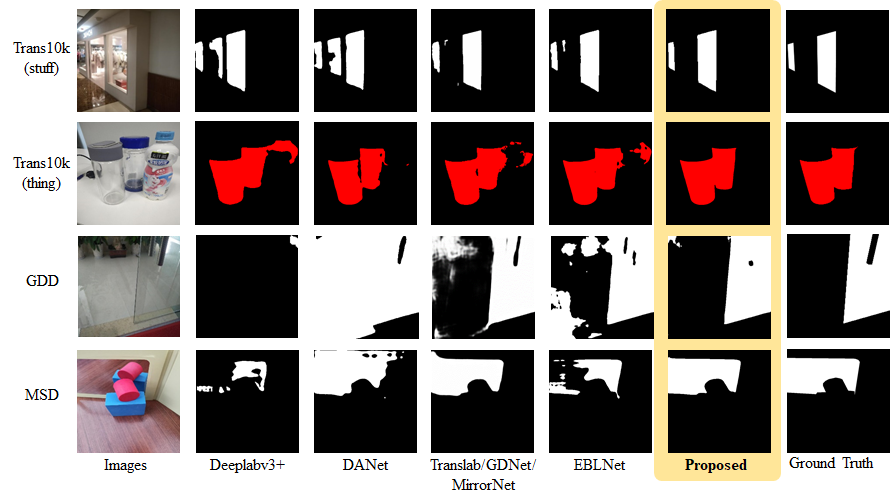}
    \vspace*{-3mm}
\caption{Sample qualitative comparison results on Trans10k, GDD, and MSD}
  \label{lastpic}
\end{figure*}
\begin{figure*}
\centering
\includegraphics[scale=0.8]{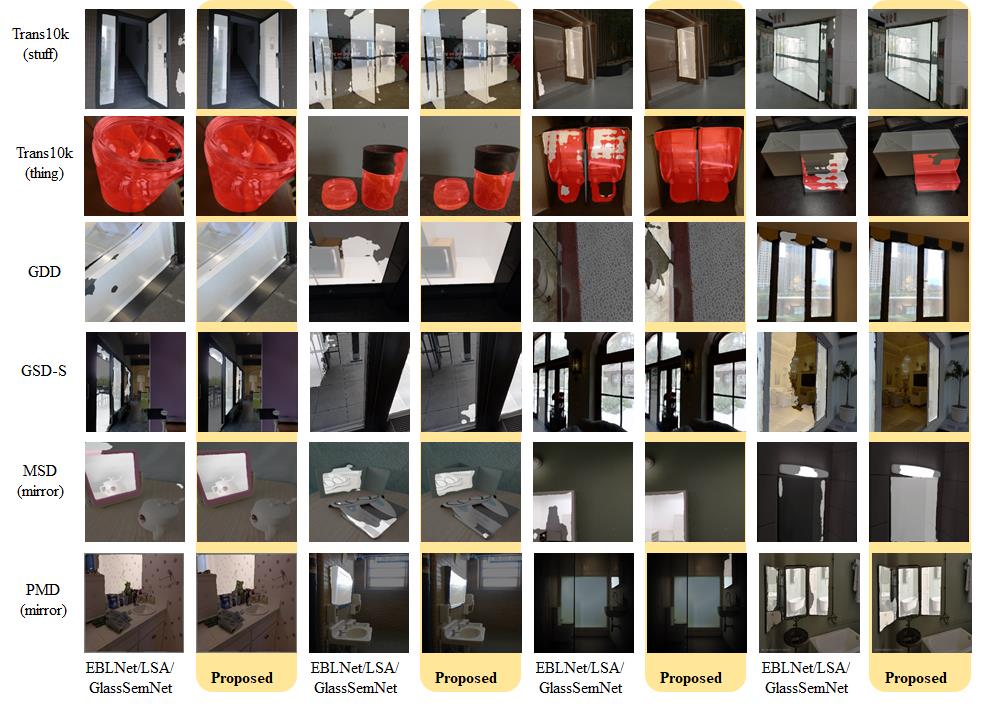}
    \vspace*{-2mm}
\caption{Additional results on Trans10k, GDD, MSD compared to EBLNet, GSD-S compared to GlassSemNet, PMD compared to LSA.}
  \label{GDDimg} 
\end{figure*}
\begin{table}\scriptsize
\begin{tabular}{lcccc} 
    \toprule
    Method & mIoU $\uparrow$ & Acc $\uparrow$ & mAE $\downarrow$ &mBER $\downarrow$ \\
    \midrule
    Deeplabv3+ \cite{chen2018reverse}(MobileNetv2) & 75.27 & 80.92 & 0.130 & 12.49\\
    HRNet \cite{wang2020deep}  & 74.56 & 75.82                 & 0.134 & 13.52\\
    BiSeNet \cite{yu2018bisenet}  & 73.93    & 77.92            & 0.140 & 13.96\\
    DenseAspp \cite{deng2018learning} & 78.11  & 81.22            & 0.114 & 12.19\\
    FCN \cite{long2015fully}  &79.67             & 83.79             & 0.108 & 10.33 \\
    RefineNet \cite{lin2019refinenet}  &66.03   &  57.97            & 0.180 & 22.22\\
    Deeplabv3+ \cite{chen2018reverse} (ResNet50) & 84.54 & 89.54      & 0.081 & 7.78\\
    Translab\cite{xie2020segmenting} (ResNet50)  &87.63 & 92.69     & 0.063 & 5.46\\
    EBLNet\cite{he2021enhanced} (ResNet50, OS16)  & 89.58 & 93.95     & 0.052 & 4.60\\
    EBLNet\cite{he2021enhanced} (ResNet50, OS8) & 90.28  & 94.71     & 0.048 &4.14\\
    \midrule
    \textbf{Proposed}(ResNet50, OS16)  & 90.09 & 94.57& 0.050 & 4.15\\
    \textbf{Proposed}(ResNet50, OS8)   & \textbf{90.79}  & \textbf{95.21} & \textbf{0.046} &\textbf{3.91}\\
    \bottomrule
    \end{tabular}
    \vspace*{-2mm}
    \caption{Experimental Comparison on Trans10k: OS indicates output stride in the backbone. Best results are indicated in bold.}
    \vspace*{-1mm}
    \label{tab:test}
\end{table}
\begin{table}\scriptsize
    \centering
    \begin{tabular}{lccccc} 
    \toprule
    Method & IoU $\uparrow$ & Acc$\uparrow$ &${F}_{\beta}$ $\uparrow$& mAE $\downarrow$ &BER $\downarrow$\\
    \midrule
    ICNet \cite{zhao2018icnet}  & 57.25 &0.694&0.710 & 0.124 &18.75\\
    DSS\cite{hou2017deeply}  & 59.11 & 0.665  &0.743  & 0.125 & 18.81\\
    RAS\cite{chen2018reverse}  & 60.48 & 0.845 &0.758     & 0.111 & 17.60\\
    BDRAR\cite{zhu2018bidirectional}  & 67.43 & 0.821 &0.792     & 0.093 & 12.41\\
    DSC\cite{hu2018direction}  & 69.71 & 0.816  &0.812    & 0.087 & 11.77\\
    MirrorNet\cite{yang2019my} (ResNext101)  &78.95 & 0.935 &0.857 & 0.065 & 6.39\\
    LSA \cite{guan2022learning} (ResNext101) &79.85 & 0.946 & 0.889    & 0.055 & 7.12\\
    EBLNet \cite{he2021enhanced} (ResNet101) & 78.84 & 0.946   &0.873  & 0.054 & 8.84\\
    EBLNet \cite{he2021enhanced} (ResNext101) & 80.33  & 0.951 &0.883    & 0.049 &8.63\\
    \midrule
    \textbf{Proposed}(ResNet101)  & 79.20 &0.949 &0.870& 0.051 & 8.47\\
    \textbf{Proposed}(ResNext101)  & \textbf{81.48}  & \textbf{0.953} & \textbf{0.899}& \textbf{0.047} &\textbf{7.00}\\
    \bottomrule
    \end{tabular}
        \vspace*{-2mm}
    \caption{Experimental Comparison on MSD}
            \vspace*{-1mm}
        \label{MSD}

\end{table}
\begin{table}\scriptsize
    \centering
    \begin{tabular}{lccccc} 
    \toprule
    Method & IoU $\uparrow$ & Acc $\uparrow$ &${F}_{\beta}$ $\uparrow$ & mAE $\downarrow$ &BER $\downarrow$ \\
    \midrule
    PSPNet \cite{zhao2017pyramid}    &84.06 &0.916 &0.906& 0.084 &8.79\\
    PointRend \cite{kirillov2020pointrend} (Deeplabv3+) & 86.51 &0.933&0.928 & 0.067& 6.50\\
    PiCANet \cite{liu2018picanet}   &83.73 &0.916&0.909 & 0.093 &8.26\\
    DSC \cite{hu2018direction} &83.56 & 0.914  &0.911   & 0.090 & 7.97\\
    BDRAR \cite{zhu2018bidirectional}  &80.01 & 0.902 &0.902    & 0.098 & 9.87\\
    MirrorNet \cite{yang2019my} (ResNext101)&85.07 & 0.918 &0.903    & 0.083 & 7.67\\
    GDNet \cite{mei2020don} (ResNext101) &87.63 & 0.939 &0.937 & 0.063 & 5.62\\
    EBLNet \cite{he2021enhanced} (ResNet101) & 88.16 & 0.941  &0.939   & 0.059 & 5.58\\
    \midrule
    \textbf{Proposed}(ResNet101) & \textbf{88.72} &\textbf{0.944} &\textbf{0.945}&\textbf{ 0.056} & \textbf{5.34}\\
    \bottomrule
    \end{tabular}
            \vspace*{-2mm}
    \caption{Experimental Comparison on GDD: GlassSemNet\cite{linexploiting} reports \textbf{90.8} of IoU, which is trained on GSD-S dataset.}
            \vspace*{-1mm}
    \label{GDD}
\end{table}
\begin{table}\scriptsize
  \centering
  \begin{tabular}{ccccccccc}
    \toprule
     &&mIoU $\uparrow$ & Acc $\uparrow$ & mAE $\downarrow$ &mBER $\downarrow$ \\
    \midrule
    &Deeplabv3+ & 85.51 & 89.97 &0.075 &6.91\\
    Study 1&+IEBAM & 90.64 &94.64  &0.046 &3.97\\
    &\scriptsize{+IEBAM+FBAM} & \textbf{91.00} & \textbf{95.64}  &\textbf{0.044} &\textbf{3.68}\\
    \midrule
    &Ex only & 90.68 &95.28  &0.046&3.81\\
   Study 2& In only  & 90.39 &94.70  &0.047&3.99\\
   & In+Ex   & \textbf{91.00} & \textbf{95.64}  &\textbf{0.044} &\textbf{3.68}\\
    \midrule
    &${\textit{L}}_{in}$/${\textit{L}}_{ex}$-bce&90.34&94.83&0.047&4.00 \\
    Study 3&${\textit{L}}_{in}$/${\textit{L}}_{ex}$-dice&90.61&94.91&0.046&3.95\\
    & ${\textit{L}}_{in}$/${\textit{L}}_{ex}$-co& \textbf{91.00} & \textbf{95.64}  &\textbf{0.044} &\textbf{3.68}\\
    \bottomrule
  \end{tabular}
          \vspace*{-2mm}
  \caption{Ablation Study 1: with/without IEBAM and FBAM, Ablation Study 2: with/without inboundary (In) and exboundary(Ex), Ablation Study 3: Effect of different loss terms on Trans10k}
          \vspace*{-1mm}
  \label{trans10kinex}
\end{table}

\begin{table}\scriptsize
  \centering
  \begin{tabular}{ccccccc}
    \toprule
     &3pix&4pix &5pix&6pix &7pix \\
    \midrule
    Things&92.96  & 92.83 &93.34 &92.96 &92.78\\
    Stuff &88.60  &88.18  &88.65 & 88.11 &88.10 \\
    MIoU  &90.78  &90.50  &91.00 &90.53 &90.44\\
    \bottomrule
  \end{tabular}
          \vspace*{-2mm}
  \caption{Ablation Study 4: Varying boundary thicknesses}
          \vspace*{-4mm}
  \label{thickness}
\end{table}

For fair comparison, hyperparameters are from \cite{he2021enhanced}. Stochastic gradient descent (SGD) optimizer, initial learning rate of 0.01 and decay \cite{liu2015parsenet} with the power of 0.9 for 16 epochs are used. Input images are augmented by random horizontal flipping and resizing with the size of 512$\times{}$512. The backbone of Trans10k is ResNet50, which uses two different output steps of 8 and 16 and pre-trained on ImageNet \cite{deng2009imagenet}, while the remaining layers of our model are randomly initialized. We utilize mIoU, mBER, mAE as segmentation metrics in ablation study, and we add ACC when comparing with other methods. 
However, in the ablation study using the Trans10k, we reduce the learning rate to half and total batch size to 8 for epoch=40 in order to conduct stable training to obtain more accurate test results.

\textbf{Comparison with State-of-the-Arts:} We compare our method with semantic segmentation methods (HRNet\cite{wang2020deep}, BiSeNet \cite{yu2018bisenet}, PSPNet\cite{gao2020highly}, DeepLabV3+\cite{chen2018reverse}, DenseASPP\cite{deng2018learning}, FCN\cite{long2015fully}, RefineNet\cite{lin2019refinenet}) and glass object segmentation methods (Translab\cite{xie2020segmenting}, EBLNet\cite{he2021enhanced}). As summarized in Table 1, proposed method achieves best scores in the four metrics (output stride 8 and 16 are set for fair comparison with EBLNet). Figure \ref{lastpic} shows compared sample results. In first image, gateway in the middle without glass and glass window are all accurately detected by our method. Second image has both non-glass and glass bottles and our method finds only glass bottles accurately.

\textbf{Ablation Study 1: with/without IEBAM and FBAM} With DeepLabV3+\cite{chen2018reverse} as baseline, we add only IEBAM ("+IEBAM") and both IEBAM and FBAM ("+IEBAM +FBAM").
In Table \ref{trans10kinex}, adding IEBAM improves mIoU by 5.13 $\%$. It shows the importance of feature extraction scheme in glass object detection. FBAM brings additional improvement of 0.36 $\%$ in mIoU. 

\textbf{Ablation Study 2: with/without inboundary (In) and
exboundary(Ex):} We conduct experiments in the FBAM network using internal ${\textit{F}}_{in}$ and external boundary features ${\textit{F}}_{ex}$ as query in order to explore the effects of internal and external boundary respectively. As shown in Table \ref{trans10kinex}, external boundary plays more dominant role than internal boundary when recognizing transparent objects. 
Most glass objects have distinct external boundaries but small number of glass objects have more distinct internal boundaries. And our algorithm FBAM deals the situation properly. 

\textbf{Ablation Study 3: Effect of different loss terms} To demonstrate the phenomenon that closer pixels to boundary at internal and external boundaries are stronger features, we test with binary cross-entropy loss (${\textit{L}}_{in}/{\textit{L}}_{Ex}-bce$), dice loss(${\textit{L}}_{in}/{\textit{L}}_{Ex}-dice$), and decoupled contour loss (${\textit{L}}_{in}/{\textit{L}}_{Ex}-co$). 
Decoupled contour loss pays more attention to the pixels near boundary region. As shown in Table \ref{trans10kinex}, our decomposed contour loss improves performance by 0.66$\%$, and 0.39$\%$ compared with Dice loss and binary cross-entropy loss respectively.

\textbf{Ablation Study 4: Varying boundary thickness:} We vary the thicknesses of internal and external boundary regions with two categories(stuff and things) of objects on Trans10k. In Table \ref{thickness}, as the thickness of internal and external boundary increase, mIoU increases up to 91.0$\%$. 

\begin{table}\scriptsize
  \centering
  \begin{tabular}{cccccccc}
    \toprule
     &&IoU$\uparrow$ & Acc$\uparrow$ &${F}_{\beta}$$\uparrow$ & mAE$\downarrow$ & BER$\downarrow$\\
    \midrule
    &Ex only  & 88.01 &0.940 &0.938 &0.059&5.63\\
    GDD&In only& 88.39 &0.943 &0.939 &0.057&5.60\\
    &In+Ex & \textbf{88.72} &\textbf{0.944}  &\textbf{0.945} &\textbf{0.056}&\textbf{5.34}\\
    \midrule
    &Ex only & \textbf{80.72} &\textbf{0.951} &\textbf{0.896} &\textbf{0.049}&\textbf{7.03}\\
    MSD&In only & 77.87 &0.944&0.876&0.056& 8.52 \\
    &In+Ex   & 79.20 &0.949&0.870&0.051&8.47\\
    \bottomrule    
  \end{tabular}
            \vspace*{-2mm}
  \caption{Ablation Study 5,6: with/without inboundary (In) and exboundary(Ex) on GDD and MSD}
            \vspace*{-4mm}
  \label{GDDabltion}
\end{table}

\subsection{Experiments on GDD and MSD}
For fair comparison, training and testing hyperparameters follow \cite{he2021enhanced}. For GDD, input images are augmented by randomly horizontal flipping and resizing and the input of the network is standardized to 416$\times{416}$ in both training and testing sets. The parameters of multi-level feature extractor are initialized by pre-trained ResNet101. Other parameters are initialized randomly. Initial learning rate is 0.003 and stochastic gradient descent (SGD) with momentum of 0.9 is used for 200 epochs.
For MSD, input images are augmented by randomly horizontal flipping and resizing and the input of the network is standardized to 384$\times{384}$. The parameters of the multi-level feature extractor are initialized by pre-trained ResNet101, ResNetX101 networks and other parameters are initialized randomly. Learning rate is 0.002 and SGD with momentum of 0.9 is used for 160 epochs.

We compare our method with the state-of-the-art semantic segmentation, shadow detection, saliency object detection, and glass segmentation methods such as Translab, GDNet, EBLNet and mirror segmentation methods such as LSA\cite{guan2022learning}, MirrorNet as shown in Figure \ref{lastpic}. Test on MSD shows that our method segments mirror region reducing the interference of the reflection inside the mirror according to the external boundary.
In Table \ref{MSD}, \ref{GDD}, our method shows best performance over prior work.
Interestingly in Table \ref{MSD}, compared to LSA that segments semantics of objects around the mirror as auxiliary features, we achieve better IoU by 1.61$\%$. 
Improvement of our method on GDD is somewhat limited. We observe that, as shown in Figure \ref{GDDimg}, GDD takes camera oblique and close up photos in order to capture the reflection phenomenon. Thus most of glass images does not show boundary. However, in such extreme cases, our method shows improvement as shown in Table \ref{GDD}.

\textbf{Ablation Study 5,6: with/without inboundary (In) and
exboundary(Ex):} 
As summarized in Table \ref{GDDabltion}, training result using only internal boundaries in GDD is improved by 0.33$\%$ in IoU by adding external boundaries and training result using only external boundaries in GDD is improved by 0.71$\%$ in IoU by adding internal boundaries. Training result using only internal boundaries in MSD is improved by 1.33$\%$ in IoU by adding external boundaries and training result using only external boundaries in MSD is improved by 0.23$\%$ in IoU by adding internal boundaries.

\subsection{Experiments on GSD, GSD-S and PMD}
Table \ref{GSD} shows performance comparison with state-of-the-art methods on GSD and GSD-S glass object datasets. Proposed method achieves best performance on the two most important metrics in semantic segmentation tasks (IoU and ${F}_{beta}$).
In Table 8, our method also shows the best performance in the three metrics on PMD. Especially proposed method achieves 2.14$\%$ of IoU improvement compared to the state-of-the-art mirror segmentation method. Directly capturing the frame information of transparent objects on the glass surface, rather than utilizing semantic information of other objects in the image during the training process, is more effective, especially in mirror segmentation tasks. 

\begin{table}\scriptsize
    \centering
    \begin{tabular}{llccccc} 
    \toprule
    &Method & IoU $\uparrow$ &${F}_{\beta}$ $\uparrow$ & mAE $\downarrow$ &BER $\downarrow$ \\
    \midrule
    &BASNet\cite{qin2019basnet}  &69.79 &0.808 &0.106& 13.54 \\
    &MINet  \cite{pang2020multi}  &77.29 &0.879 &0.077& 9.54 \\
    
    &SINet \cite{fan2020camouflaged} &77.04 &0.875&0.077 & 9.25\\
    GSD &GDNet \cite{mei2020don}  &79.01 &0.869&0.069 &7.72  \\
    &TransLab\cite{xie2020segmenting} &74.05 & 0.837  &0.088  & 11.35\\
   & GlassNet \cite{lin2021rich}    &83.64 & 0.903 &0.055    & 6.12 \\
   & GlassSemNet\cite{linexploiting} &85.60 &0.920 &\textbf{0.044} &\textbf{5.60}\\
    \midrule
   & \textbf{Proposed} &\textbf{86.12} & \textbf{0.926} &0.049 & 6.27\\
    \midrule
        \midrule
   & PSPNet \cite{zhao2017pyramid}    &56.0 &0.679 &0.093& 13.40 \\
   & PSANet &55.0 &0.656 &0.104& 12.61 \\
   & SETR \cite{zhu2018bidirectional}  &56.7 &0.679    & 0.086 & 13.25\\
   & Segmenter \cite{yang2019my} &53.6 & 0.645 &0.101    & 14.02\\
   & Swin \cite{yang2019my}&59.6 & 0.702 &0.082    & 11.34\\
  GSD-S & ViT \cite{yang2019my} &56.2 &0.693    & 0.087 & 14.72\\
   & SegFormer \cite{yang2019my} &54.7 & 0.683 &0.094    & 15.15\\
   & Twins \cite{yang2019my} &59.0 & 0.703 &0.084    & 12.43\\
  &  GDNet \cite{mei2020don}&52.9 & 0.642 &0.101   & 18.17 \\
   & GlassNet \cite{lin2021rich}  &72.1 & 0.821 &0.061 & 10.02 \\
   & GlassSemNet \cite{linexploiting} &75.3&0.860 &\textbf{0.035}&\textbf{9.26}\\
    \midrule
   & \textbf{Proposed}& \textbf{77.92} &\textbf{0.874} & 0.040& 9.93 \\
    \bottomrule
    \end{tabular}
    \vspace*{-2mm}
    \caption{Experimental Comparison on GSD and GSD-S 
    }
    \vspace*{-2mm}
    \label{GSD}
\end{table}

\begin{table}\scriptsize
    \centering
    \begin{tabular}{lccccc} 
    \toprule
    Method & IoU $\uparrow$ &${F}_{\beta}$ $\uparrow$ & mAE $\downarrow$ &Acc $\uparrow$ \\
    \midrule
    CPNET \cite{yu2020context}  &56.36 &0.734 &0.051& 94.85 \\
    GloRe \cite{chen2019graph} &61.25 &0.774 &0.044 & 95.61\\
    PSPNet  \cite{zhao2017pyramid} &60.44 &0.806    &0.039 &96.13  \\
    MirrorNet \cite{yang2019my} &62.50 & 0.778  &0.041  & 96.27\\
    PMD-Net \cite{lin2020progressive}  &62.40 & 0.827 &0.055    & 96.80 \\
    LSA  \cite{guan2022learning}&66.84 & \textbf{0.844} &0.049  & 96.82\\
    \midrule
    \textbf{Proposed} & \textbf{68.90} & 0.816 &\textbf{0.029} & \textbf{97.10}\\
    \bottomrule
    \end{tabular}
        \vspace*{-2mm}
    \caption{Experimental Comparison on PMD
    }    \vspace*{-4mm}
    \label{PMD}
\end{table}

\subsection{Conclusion}
We propose IEBAM and FBAM for glass object segmentation. Extensive evaluations on six public datasets are conducted showing outperforming glass object segmentation performance over prior work. We observe that the external frame region of glass objects plays a critical role than any other semantic segmentation information of objects avoiding overfitting of a model to the dataset.




{\small
\bibliographystyle{ieee_fullname}
\bibliography{egbib}
}

\end{document}